\documentclass[conference,letterpaper]{IEEEtran}
\IEEEoverridecommandlockouts
\usepackage[letterpaper,top=0.85in,bottom=1.1in,left=1.05in,right=1.05in]{geometry}
\usepackage{amsmath,amsfonts}
\usepackage{algorithmic}
\usepackage{algorithm}
\usepackage{array}
\usepackage[caption=false,font=normalsize,labelfont=sf,textfont=sf]{subfig}
\usepackage{textcomp}
\usepackage{stfloats}
\usepackage{url}
\usepackage{verbatim}
\usepackage{graphicx}
\usepackage{cite}
\usepackage{microtype}
\begin{document}

\title{FedOrbit: Adaptive Personalized Federated Learning for Non-IID LEO Satellite Constellations}
\author{
\IEEEauthorblockN{
  1\textsuperscript{st} Satwat Bashir,\quad
  2\textsuperscript{nd} Tasos Dagiuklas,\quad
  3\textsuperscript{rd} Muddesar Iqbal
}
\IEEEauthorblockA{
  \textit{School of Engineering} \\
  \textit{London South Bank University}, London, UK \\
  \{bashis11, tdagiuklas, m.iqbal\}@lsbu.ac.uk
}
\thanks{This work has received funding from the European Commission under the Horizon Europe MSCA programme (HORIZON-MSCA-2024-SE-01-01), Grant Agreement No. 101236523 (AeroNet project).}
}

\maketitle

\begin{abstract}
Federated learning (FL) in Low Earth Orbit (LEO) satellite
constellations is affected by non-IID data and irregular
ground-station visibility, both driven by orbital geometry.
Global aggregation performs poorly when orbit-level class
distributions are disjoint, while strong personalisation can be
excessive when these distributions overlap. We
present FedOrbit, which combines continuous orbit-level
training over inter-satellite links, class-aware hierarchical
aggregation, quality-weighted feature aggregation with
return-rate dampening, and adaptive feature decomposition
based on inter-orbit class similarity. Across three
remote-sensing benchmarks and two non-IID partitions, FedOrbit
achieves the highest accuracy in five of six settings and is
within $0.9$ percentage points of the best result in the sixth.
The gains over the strongest baseline reach $16.1$ percentage
points under Dirichlet partitioning and $8.6$ under pathological
partitioning, with
the smallest per-orbit accuracy spread in five of six settings.
\end{abstract}

\begin{IEEEkeywords}
Federated learning, LEO satellite constellations, personalised aggregation, non-IID data, non-terrestrial networks.
\end{IEEEkeywords}

\section{Introduction}
\label{sec:intro}

Low Earth Orbit (LEO) satellite constellations are now an
important platform for global Earth
observation~\cite{kodheli2021survey,delportillo2019comparison},
and on-board processing makes it possible to train and run
machine learning models directly on the
satellite~\cite{giuffrida2020cloudscout,mateogarcia2021flood}.
Transmitting raw observation data to a ground station for
centralised training is impractical given the intermittent
nature of satellite-to-ground links, data privacy
requirements, and bandwidth constraints. Federated learning
(FL)~\cite{mcmahan2017fedavg} reduces the need for raw-data
transfer: each
satellite trains locally on its own observations and exchanges
only model parameters with a ground station (GS).

FL in LEO constellations differs fundamentally from the
terrestrial setting. In a Walker Delta constellation, each
orbit follows a distinct ground track and observes a different
geographic region, producing orbit-level class distributions
that are inherently
non-IID~\cite{matthiesen2023flconstellations,leyva2020leo}.
Orbit-to-GS visibility is governed by the same orbital
geometry, producing imbalanced participation. In the
constellation considered in this work, approximately $47\%$ of
training rounds have
no orbit visible to the GS, and the least-visited orbit
participates in fewer than one-third as many rounds as the
most-visited orbit~\cite{razmi2022ground}. As the class
distribution of an orbit is linked to its ground track and its
visibility schedule is determined by orbital geometry, class
contribution and timing are systematically correlated rather
than independent. This correlation between data
heterogeneity and participation is the defining property of
the LEO FL setting.

This paper presents FedOrbit, a personalised FL framework
that addresses the correlation between data and participation
in LEO constellations. The contributions are:
(i)~identification of two failure modes induced by this correlation,
namely global-model collapse on pathological partitioning and
over-personalisation on Dirichlet partitioning;
(ii)~\emph{continuous orbit-level training} over
inter-satellite links, so every orbit refines the model in
every round;
(iii)~\emph{class-aware hierarchical aggregation}, routing
classifier updates by per-class data ownership;
(iv)~\emph{quality-weighted feature aggregation with
return-rate dampening}, limiting the influence of an orbit
returning from a long absence;
(v)~\emph{adaptive feature decomposition}, blending each
orbit's personal feature extractor toward the global one at a
rate set by inter-orbit class similarity.
We evaluate FedOrbit on three remote-sensing classification
benchmarks under both pathological and Dirichlet
($\alpha = 0.5$) partitions, with the largest gains on the
partition where existing methods perform worst.

\section{Related Work}
\label{sec:related}

\textbf{FL under non-IID data.}
FedAvg~\cite{mcmahan2017fedavg} degrades under heterogeneous
distributions~\cite{li2022niid};
FedProx~\cite{li2020fedprox},
SCAFFOLD~\cite{karimireddy2020scaffold}, and
FedNova~\cite{wang2020fednova} use proximal regularisation,
control variates, or local-step normalisation. These methods
produce a single global model but do not model the coupling
between local data and visibility-dependent participation.

\textbf{Personalised FL.} Ditto~\cite{li2021ditto} adds an
$\ell_2$ penalty toward the global model;
APFL~\cite{deng2020apfl} interpolates local and global models;
other methods use regularised inner
loops~\cite{dinh2020pfedme}, meta-learned
initialisations~\cite{fallah2020perfedavg}, or split
representations~\cite{collins2021fedrep}. These methods do not
account for bias in the global reference model caused by
visibility-dependent participation.

\textbf{FL in non-terrestrial networks.} Prior LEO FL work has
addressed scheduling under irregular
visibility~\cite{razmi2022ground}, semi-asynchronous
aggregation~\cite{so2022fedspace}, aerial parameter
servers~\cite{elmahallawy2022fedhap,elmahallawy2022asyncfleo},
decentralised intra-orbit aggregation~\cite{zhai2024fedleo},
over-the-air computation~\cite{huang2024asyncfl}, system
heterogeneity~\cite{lin2025fedsn}, and decentralised
personalisation~\cite{zhao2024alanine}. The common focus is
communication architecture, convergence speed, and scheduling.
To our knowledge, prior work does not jointly address
orbit-level non-IID data and visibility-dependent participation
or adapt personalisation to inter-orbit similarity.

\section{System Model}
\label{sec:system}

\subsection{LEO Constellation and Visibility Model}
\label{sec:constellation}

We consider a Walker Delta
constellation with parameters
following prior FL-over-LEO
work~\cite{razmi2022ground}: $L = 5$ orbits, $N_l = 4$
satellites per orbit (20 satellites total), altitude
$h = 550$\,km, inclination $i = 53^{\circ}$, GS at latitude
$\phi_g = 51^{\circ}$\,N, minimum elevation
$\alpha_e = 10^{\circ}$, and round duration
$\Delta t = 5$\,min. The GS acts as the parameter server.

Each satellite follows a circular Keplerian orbit with mean
motion $n_o = \sqrt{\mu/(R_E + h)^3}$, where
$\mu = 3.986 \times 10^5$\,km$^3$/s$^2$ is Earth's
gravitational parameter and $R_E = 6371$\,km is Earth's mean
radius. The right ascension of the ascending node (RAAN) of
orbit $l$ is $\Omega_l = 2\pi l / L$ for $l = 0, \ldots, L-1$,
giving equally spaced orbits. The GS position in the
Earth-centred inertial (ECI) frame rotates with Earth at
angular velocity $\omega_E = 7.292 \times 10^{-5}$\,rad/s.
Satellite $k$ in orbit $l$ is visible from the GS when
\begin{equation}
\label{eq:elevation}
\varepsilon_k(t) = \arcsin\!\left(
  \frac{(\mathbf{r}_k(t) - \mathbf{r}_g(t)) \cdot
        \hat{\mathbf{r}}_g(t)}
       {\|\mathbf{r}_k(t) - \mathbf{r}_g(t)\|}
\right) \geq \alpha_e\,,
\end{equation}
where $\mathbf{r}_k(t)$ and $\mathbf{r}_g(t)$ are the
satellite and GS positions in the ECI frame and
$\hat{\mathbf{r}}_g(t)$ is the local zenith unit vector at the
GS. An orbit is treated as visible at round $t$ if at least
one of its satellites satisfies~\eqref{eq:elevation}, since
the orbit-level model is what is uploaded to the GS. A binary
matrix $\mathbf{V} \in \{0,1\}^{T \times L}$ records
visibility, with $V_{t,l} = 1$ if orbit $l$ is visible at
round $t$ and $0$ otherwise; the set of visible orbits at
round $t$ is $\mathcal{V}(t) = \{l : V_{t,l} = 1\}$. All
evaluated methods use the same $\mathbf{V}$.

For $T = 400$, approximately $47\%$ of rounds have no visible
orbit, the per-orbit visibility shares are
$15.2\%, 10.2\%, 8.5\%, 5.2\%$, and $14.0\%$, and the longest
contiguous interval with no visible orbit is
$\tau^{\max} = 191$ rounds.

\subsection{Hierarchical Federated Learning Architecture}
\label{sec:fl_arch}

\begin{figure}[!htbp]
\centering
\includegraphics[width=0.95\columnwidth]{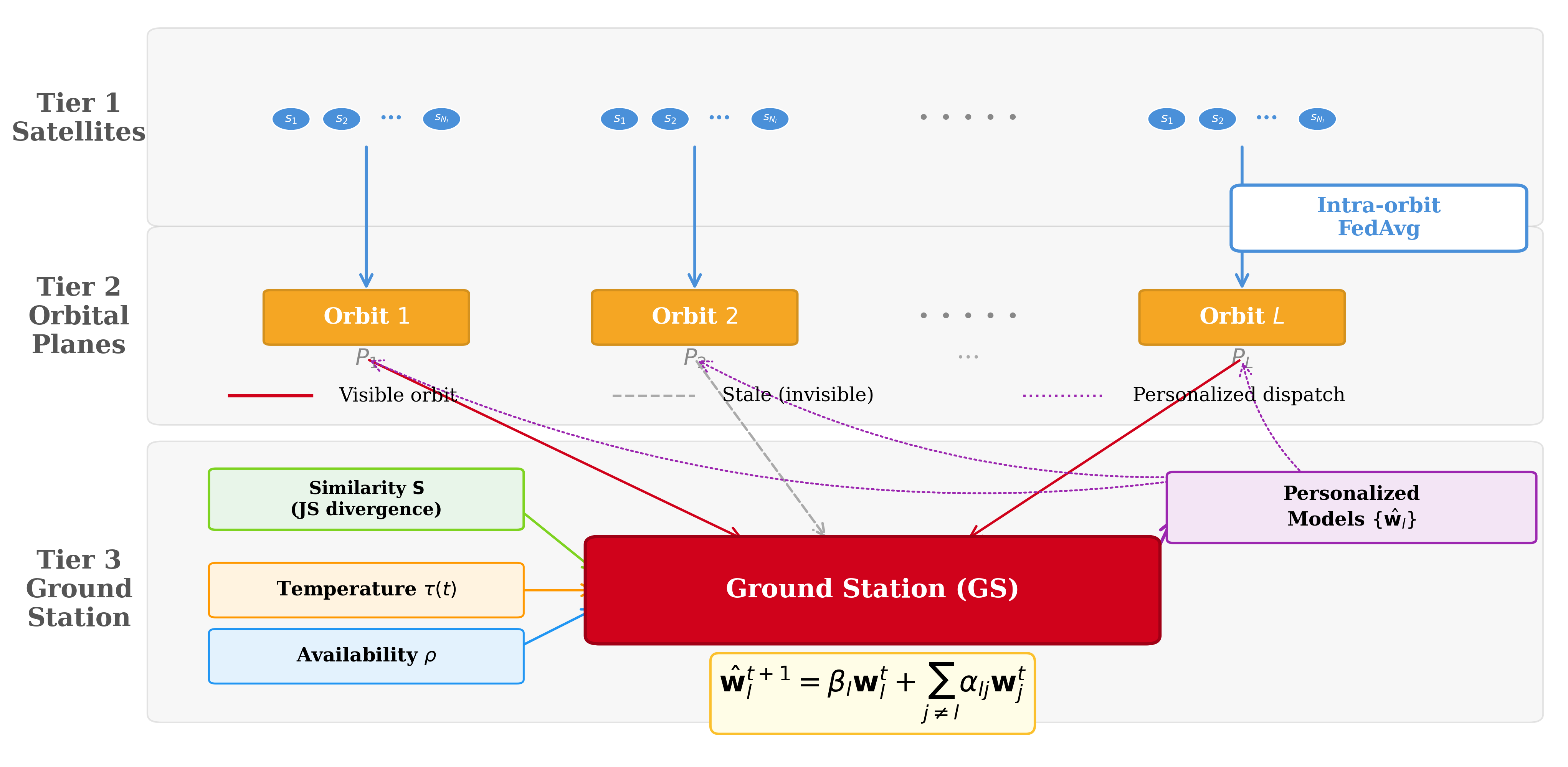}
\caption{FedOrbit architecture. Satellites
train locally, aggregate within each orbit over ISLs, and
downlink the orbit-level model to the GS during visibility
windows.}
\label{fig:system}
\end{figure}
\enlargethispage{-0.2in}

The architecture has satellite, orbit, and GS tiers
(Fig.~\ref{fig:system}) and is shared by all methods. The
following FedAvg-based procedure is used by the baselines.
FedOrbit modifies both aggregation tiers and trains
non-visible orbits over inter-satellite links, as described in
Section~\ref{sec:method}.

\textit{Local training.} Satellite $k$ holds a dataset
$\mathcal{D}_k$ of size $n_k$ with local objective
$F_k(\mathbf{w}) = (1/n_k)\sum_{x \in \mathcal{D}_k}
f(x; \mathbf{w})$, and runs $E$ epochs of mini-batch SGD with
learning rate $\eta_t = \eta_0 \gamma^{t}$, producing
\begin{equation}
\label{eq:local_sgd}
\mathbf{w}_k^{t} = \texttt{LocalSGD}\bigl(
  \mathbf{w}^{t},\, \mathcal{D}_k,\, E,\, \eta_t \bigr).
\end{equation}

\textit{Intra-orbit aggregation.} Satellites in orbit $l$
combine their updates over short-range ISLs by sample-weighted
averaging:
\begin{equation}
\label{eq:intra_orbit}
\mathbf{w}_l^{t} = \sum_{k \in \mathcal{S}_l}
  \frac{n_k}{\sum_{j \in \mathcal{S}_l} n_j}\,
  \mathbf{w}_k^{t},
\end{equation}
where $\mathcal{S}_l$ is the set of satellites in orbit $l$.

\textit{Inter-orbit aggregation at the GS.} Baselines apply
FedAvg~\cite{mcmahan2017fedavg} across the visible orbits:
\begin{equation}
\label{eq:fedavg_global}
\mathbf{w}^{t+1} = \sum_{l \in \mathcal{V}(t)}
  \frac{n_l}{\sum_{j \in \mathcal{V}(t)} n_j}\,
  \mathbf{w}_l^{t},
\end{equation}
with $n_l = \sum_{k \in \mathcal{S}_l} n_k$. FedProx, APFL,
and Ditto use the same weighting and differ from FedAvg only
in their local objectives.

\subsection{Non-IID Data Distribution Model}
\label{sec:noniid}

Each orbit collects imagery along its ground track, producing
a class distribution $P_l$ that reflects the terrain it
overflies. Because satellites in the same orbit share the same
ground track, heterogeneity is orbit-specific rather than
satellite-specific. Two partitioning methods are considered to
represent contrasting levels of inter-orbit overlap:

\textit{Pathological partitioning.} Each orbit is assigned a
disjoint subset of classes, modelling
orbits that observe disjoint land-cover categories. Within
each orbit, satellite allocations follow a
Dirichlet$(\alpha = 0.5)$ split.

\textit{Dirichlet partitioning ($\alpha = 0.5$).} Inter-orbit
class proportions are drawn from Dirichlet$(0.5)$, giving
overlapping but skewed orbit distributions. Intra-orbit
allocation also follows Dirichlet$(0.5)$. This models the
partial overlap that arises when orbits cover geographically
proximate but distinct regions.

\section{Problem Formulation}
\label{sec:problem}

In the Walker Delta constellation of
Section~\ref{sec:system}, the class distribution $P_l$ is linked
to the ground track of orbit $l$, while the visibility schedule
$\{V_{t,l}\}_{t=1}^{T}$ is determined by orbital geometry
through~\eqref{eq:elevation}. The standard FL assumption that
client participation is independent of local
data~\cite{mcmahan2017fedavg,li2020fedprox,karimireddy2020scaffold}
therefore does not hold, and the global model cannot serve as
an unbiased reference for personalisation.

Let $\{\widetilde{\mathbf{w}}_l\}_{l=1}^{L}$ denote the
orbit-level inference models, $F_l$ the expected loss on
$P_l$, and $\mathrm{Acc}_l$ the corresponding accuracy.
FedOrbit is guided by the following fairness-constrained objective:
\begin{equation}
\label{eq:objective}
\min_{\{\widetilde{\mathbf{w}}_l\}}
  \tfrac{1}{L} \sum_{l=1}^{L} F_l(\widetilde{\mathbf{w}}_l)
  \quad \text{s.t.} \quad
  \max_{l} \mathrm{Acc}_l - \min_{l} \mathrm{Acc}_l \leq \delta.
\end{equation}
Equation~\eqref{eq:objective} expresses the design objective;
the fairness term is not explicitly enforced during
optimisation. The empirical spread is reported in
Section~\ref{sec:results}.

The formulation imposes four design requirements:
(\textbf{R1})~reduce contribution imbalance caused by
$\mathbf{V}$; (\textbf{R2})~weight classifier updates by class
ownership; (\textbf{R3})~adapt personalisation to inter-orbit
heterogeneity without manual retuning; and (\textbf{R4})~keep
GS-uplink traffic comparable to FedAvg, with additional work
performed on-board or within an orbit.

\section{Proposed Method: FedOrbit}
\label{sec:method}

FedOrbit modifies on-board scheduling, intra-orbit and
inter-orbit aggregation, and per-orbit inference. The model is
split into a feature extractor and a
classifier,
$\mathbf{w} = [\mathbf{w}^{\mathrm{f}}, \mathbf{w}^{\mathrm{c}}]$,
since the feature extractor encodes content shared across the
constellation while each classifier row is tied to a specific
class.

\subsection{Continuous Orbit-Level Training with ISL Relay}
\label{sec:method_continuous}

Under standard FL, the least-visited orbit has almost no
influence on the trained model because it participates in only
a small fraction of rounds. FedOrbit trains every orbit in
every round, using two procedures depending on visibility.

\textit{Visible orbits.} For $l \in \mathcal{V}(t)$, the
number of local epochs grows with the length of the preceding
non-visible interval:
\begin{equation}
\label{eq:m_adaptive_epochs}
E_l(t) = \min\!\bigl(E_{\max},\;\;
  E_{\mathrm{base}} + s_e\, \tau_l(t)\bigr),
\end{equation}
where $\tau_l(t)$ is the number of consecutive non-visible
rounds prior to $t$, $s_e \in (0,1)$ scales the catch-up, and
$E_{\max}$ caps the load. Each of the
$R_{\mathrm{intra}}$ intra-orbit rounds uses $E_l(t)$ local
epochs and is followed by an intra-orbit aggregation step
(Section~\ref{sec:method_classaware}).

\textit{Non-visible orbits (ISL relay).} For
$l \notin \mathcal{V}(t)$, each satellite in the orbit runs
$E_{\mathrm{dark}}$ epochs of local SGD on the orbit-level
model state, after which an intra-orbit aggregation produces
an updated orbit model. This step uses only intra-orbit ISLs
and does not consume the GS uplink. $E_{\mathrm{dark}}$ is
kept small (two epochs) so non-visible refinements do not
overwrite the heavier visible updates.

\subsection{Class-Aware Hierarchical Aggregation}
\label{sec:method_classaware}

When a class is concentrated on a rarely visible orbit,
ordinary averaging dilutes the corresponding classifier row
with updates from orbits that do not hold the class. FedOrbit
weights aggregation by per-class data ownership at both the
intra-orbit and inter-orbit tiers.

\textit{Class strength and class affinity.} For satellite
$k \in \mathcal{S}_l$ and class $c \in \{1, \dots, C\}$,
\begin{equation}
\label{eq:m_strength}
s_{k,c} = \bigl|\{x \in \mathcal{D}_k : y(x) = c\}\bigr|,
\end{equation}
and the class affinity of orbit $l$ for class $c$ is
\begin{equation}
\label{eq:m_affinity}
a_{l,c} =
  \frac{\sum_{k \in \mathcal{S}_l} s_{k,c}}
       {\sum_{j=1}^{L} \sum_{k \in \mathcal{S}_j} s_{k,c}}.
\end{equation}
Both quantities are computed once at initialisation and remain
fixed throughout training.

\textit{Intra-orbit aggregation.} The classifier row for
class $c$ is averaged across the orbit's satellites by class
strength, and the feature extractor by the magnitudes of the
per-satellite updates
$\Delta\mathbf{w}_k^{\mathrm{f}} = \mathbf{w}_k^{\mathrm{f}} - \mathbf{w}^{\mathrm{f}}$:
\begin{align}
\mathbf{w}_l^{\mathrm{c}}[c,:] &= \sum_{k \in \mathcal{S}_l}
  \frac{s_{k,c}}{\sum_{j \in \mathcal{S}_l} s_{j,c}}\,
  \mathbf{w}_k^{\mathrm{c}}[c,:], \label{eq:m_intra_classifier} \\
\mathbf{w}_l^{\mathrm{f}} &= \sum_{k \in \mathcal{S}_l}
  \frac{\bigl|\Delta\mathbf{w}_k^{\mathrm{f}}\bigr|}
       {\sum_{j \in \mathcal{S}_l} \bigl|\Delta\mathbf{w}_j^{\mathrm{f}}\bigr|}\;
  \mathbf{w}_k^{\mathrm{f}}, \label{eq:m_intra_features}
\end{align}
where $|\cdot|$ is the elementwise absolute value, and the
ratio in~\eqref{eq:m_intra_features} is elementwise.

\textit{Inter-orbit aggregation of the classifier.} At the
GS, the classifier row for class $c$ is averaged across all
$L$ orbits with weights given by class affinity times a
staleness factor:
\begin{equation}
\label{eq:m_gs_classifier}
\mathbf{w}^{\mathrm{c}}[c,:] \leftarrow
  \frac{\sum_{l=1}^{L} a_{l,c}\, \rho^{\sigma_l(t)}\, \mathbf{w}_l^{\mathrm{c}}[c,:]}
       {\sum_{l=1}^{L} a_{l,c}\, \rho^{\sigma_l(t)}},
\end{equation}
with $\rho \in (0,1]$ and $\sigma_l(t)$ the number of rounds
since orbit $l$ last updated its classifier; $\sigma_l(t)$ is
small because every orbit trains in every round.

\subsection{Quality-Weighted Feature Aggregation and Return-Rate Dampening}
\label{sec:method_quality}

Continuous training produces orbit-level updates with
different training budgets: a visible orbit
returning from a long non-visible interval performs up to
$E_{\max} \cdot R_{\mathrm{intra}}$ epochs, while a non-visible
orbit performs only $E_{\mathrm{dark}}$. FedOrbit weights
inter-orbit feature aggregation by the amount of training each
orbit performed:
\begin{subequations}
\label{eq:m_gs_features}
\begin{align}
\bar{\mathbf{w}}^{\mathrm{f}}(t) &=
  \frac{\sum_{l=1}^{L} q_l(t)\, \mathbf{w}_l^{\mathrm{f}}}
       {\sum_{l=1}^{L} q_l(t)},
       \label{eq:m_gs_average} \\
q_l(t) &=
  \begin{cases}
    E_l(t)\, R_{\mathrm{intra}}, & l \in \mathcal{V}(t), \\
    E_{\mathrm{dark}}, & l \notin \mathcal{V}(t).
  \end{cases}
  \label{eq:m_quality_weight}
\end{align}
\end{subequations}
Under the configuration used in our experiments
($E_{\mathrm{base}} = 5$, $E_{\max} = 10$,
$R_{\mathrm{intra}} = 2$, $E_{\mathrm{dark}} = 2$), a visible
orbit contributes a weight of $10$ to $20$ and a non-visible
orbit a weight of $2$.

A returning orbit can train at the global learning rate for up
to $E_{\max}$ epochs in each intra-orbit round, producing a
feature shift that other
orbits' classifier rows have not adapted to. To attenuate
this, FedOrbit dampens the local rate of any orbit re-entering
visibility in proportion to the length of the non-visible
interval that just ended:
\begin{equation}
\label{eq:m_dampen}
\eta_l(t) =
  \frac{\eta_t}{1 + \kappa\, \tau_l(t)/\tau^{\max}},
\end{equation}
where $\tau^{\max} = 191$ rounds is the longest non-visible
interval and $\kappa \geq 0$ is a hyperparameter. For
$\kappa = 0.5$, an orbit returning after the longest interval uses
$\eta_l(t) = \eta_t / 1.5$, two thirds of the global value;
an orbit visible in the previous round ($\tau_l(t) = 0$) is
not dampened. The rule is applied only on the first round
after re-entry.

\subsection{Adaptive Feature Decomposition}
\label{sec:method_adaptive}

The aggregation rules above produce a single global
$[\bar{\mathbf{w}}^{\mathrm{f}}(t), \mathbf{w}^{\mathrm{c}}(t)]$.
A single global model is appropriate when class distributions
overlap and suboptimal when they are disjoint. FedOrbit
adapts between these two cases using a single coefficient
computed from the data.

For each orbit $l$, let $\mathbf{p}_l \in \Delta^{C-1}$ be the
normalised class histogram aggregated over its satellites,
$p_{l,c} = \sum_{k} s_{k,c} / \sum_{c',k} s_{k,c'}$. The
Jensen-Shannon distance between orbits $i$ and $j$ is
\begin{equation}
\label{eq:m_jsd}
\mathrm{JS}(\mathbf{p}_i, \mathbf{p}_j) =
  \sqrt{\tfrac{1}{2} D_{\mathrm{KL}}(\mathbf{p}_i \| \mathbf{m})
        + \tfrac{1}{2} D_{\mathrm{KL}}(\mathbf{p}_j \| \mathbf{m})},
\end{equation}
with $\mathbf{m} = \tfrac{1}{2}(\mathbf{p}_i + \mathbf{p}_j)$
and $D_{\mathrm{KL}}$ using $\log_2$, so
$\mathrm{JS} \in [0, 1]$. The inter-orbit similarity
coefficient is
\begin{equation}
\label{eq:m_beta}
\beta = \frac{2}{L(L-1)} \sum_{i<j}
  \bigl(1 - \mathrm{JS}(\mathbf{p}_i, \mathbf{p}_j)\bigr).
\end{equation}
The coefficient is computed once and fixed.

Each orbit $l$ maintains a personal feature extractor
$\mathbf{w}_l^{\mathrm{f,p}}$, initialised at
$\mathbf{w}_l^{\mathrm{f,p}}(0) = \mathbf{w}^{\mathrm{f}}(0)$
and updated by an exponential blend toward the global
quality-weighted average:
\begin{equation}
\label{eq:m_blend}
\mathbf{w}_l^{\mathrm{f,p}}(t+1) =
  (1 - \beta)\, \mathbf{w}_l^{\mathrm{f,p}}(t)
  + \beta\, \bar{\mathbf{w}}^{\mathrm{f}}(t).
\end{equation}
The same $\beta$ is used for all orbits and rounds. The
orbit's inference model is the pair of its personal feature
extractor and the global classifier:
\begin{equation}
\label{eq:m_plane_model}
\widetilde{\mathbf{w}}_l(t) =
  \bigl[\mathbf{w}_l^{\mathrm{f,p}}(t),\;
        \mathbf{w}^{\mathrm{c}}(t)\bigr].
\end{equation}
A larger $\beta$ moves the personal extractor toward the
global one faster; a smaller $\beta$ preserves orbit-specific
information. Both partitions are handled by the same procedure
with the same hyperparameters.

\subsection{Communication and Computation Cost}
\label{sec:method_cost}

\begin{algorithm}[t]
\caption{FedOrbit: one communication round.}
\label{alg:fedorbit}
\begin{algorithmic}[1]
\REQUIRE $\mathbf{V}$; $\{s_{k,c}\}$, $\{a_{l,c}\}$ from \eqref{eq:m_strength}, \eqref{eq:m_affinity}; $\beta$ from \eqref{eq:m_beta}; $\tau^{\max}$
\STATE State: $\mathbf{w}^{\mathrm{c}}(t)$, $\bar{\mathbf{w}}^{\mathrm{f}}(t)$, $\{\mathbf{w}_l^{\mathrm{f,p}}(t)\}$, $\{\tau_l, \sigma_l\}$; compute $\mathcal{V}(t)$
\FORALL{$l \in \mathcal{V}(t)$ \hfill $\triangleright$ visible}
  \STATE Set $E_l(t)$ via \eqref{eq:m_adaptive_epochs}; $\eta_l(t)$ via \eqref{eq:m_dampen}; init from \eqref{eq:m_plane_model}
  \FOR{$r = 1, \dots, R_{\mathrm{intra}}$}
    \STATE Each $k \in \mathcal{S}_l$ runs $E_l(t)$ epochs at $\eta_l(t)$; aggregate via \eqref{eq:m_intra_classifier}, \eqref{eq:m_intra_features}
  \ENDFOR
  \STATE $q_l = E_l(t) R_{\mathrm{intra}}$; reset $\tau_l, \sigma_l$
\ENDFOR
\FORALL{$l \notin \mathcal{V}(t)$ \hfill $\triangleright$ non-visible}
  \STATE Init from \eqref{eq:m_plane_model}; each $k$ runs $E_{\mathrm{dark}}$ epochs at $\eta_t$; aggregate via \eqref{eq:m_intra_classifier},\eqref{eq:m_intra_features}
  \STATE $q_l = E_{\mathrm{dark}}$; $\tau_l, \sigma_l \mathrel{+}= 1$
\ENDFOR
\STATE GS: $\bar{\mathbf{w}}^{\mathrm{f}}(t) \leftarrow$ \eqref{eq:m_gs_features}; $\mathbf{w}^{\mathrm{c}}(t+1) \leftarrow$ \eqref{eq:m_gs_classifier}; $\mathbf{w}_l^{\mathrm{f,p}}(t+1) \leftarrow$ \eqref{eq:m_blend}
\STATE \textbf{return} $\{\widetilde{\mathbf{w}}_l(t+1)\}$ via \eqref{eq:m_plane_model}
\end{algorithmic}
\end{algorithm}

Each visible orbit uploads one orbit-level model to the GS, as
in FedAvg. FedOrbit adds intra-orbit ISL exchanges and local
training during non-visible periods. Both operations are
performed in parallel across orbits. For a model with $P$ parameters, an
FP32 model requires approximately $4P$ bytes; training also
requires memory for gradients, momentum, and activations. The
sub-million-parameter LeNet models used here therefore require
tens of megabytes of working memory. A practical deployment
would require an on-board multicore processor supported by an
FPGA, GPU, or AI accelerator.

At the GS, aggregation and personal-model storage grow
linearly with $L$, while the one-time similarity calculation
in~\eqref{eq:m_beta} has complexity $\mathcal{O}(L^2C)$. The
visibility matrix $\mathbf{V}$ supports time-varying
connectivity, but larger and dynamically reconfigured
constellations have not been evaluated. These settings and
hardware-based validation are left for future work.

\section{Experimental Setup}
\label{sec:experiments}

\textit{Constellation and visibility.} All experiments use
the Walker Delta configuration of
Section~\ref{sec:constellation}. The visibility matrix
$\mathbf{V}$ is precomputed via \eqref{eq:elevation} and used
by every method. Each simulation uses $T = 200$ communication
rounds for EuroSAT and So2Sat~LCZ42 and $T = 400$ for
RESISC45.

\textit{Datasets.}
EuroSAT~\cite{helber2019eurosat}: Sentinel-2 land use,
$10$ classes, $64{\times}64$ RGB, $27{,}000$ samples.
So2Sat LCZ42~\cite{zhu2020so2sat}: Sentinel-2 Local
Climate Zone, $10$ urban classes, $32{\times}32$, $10$ bands,
approximately $352{,}000$ training and $24{,}000$ test samples.
NWPU-RESISC45~\cite{cheng2017resisc45}: aerial scenes,
$45$ classes, $32{\times}32$ RGB, $700$ images per class,
$80/20$ split.

\textit{Non-IID partitioning.} Both partitions in
Section~\ref{sec:noniid} are evaluated. The test samples are
distributed among orbits using the corresponding partitioning
rule.

\textit{Model and training.} The on-board model is LeNet (two
$5{\times}5$ convolutional layers of width $6$ and $16$ with
$2{\times}2$ max-pooling, followed by fully connected layers
\texttt{fc1}, \texttt{fc2}, \texttt{fc3} of widths $120$, $84$,
$C$). Input channels are $3$ for EuroSAT and RESISC45 and $10$
for So2Sat. The feature/classifier split is between
\texttt{fc1} and \texttt{fc2}, and the class-affinity
weighting~\eqref{eq:m_gs_classifier} is applied to the rows of
\texttt{fc3}. Training hyperparameters
(Table~\ref{tab:hparams}) are identical across methods within
each dataset.

\begin{table}[t]
\centering
\caption{Training Hyperparameters}
\label{tab:hparams}
\renewcommand{\arraystretch}{1.05}
\setlength{\tabcolsep}{3pt}
\begin{tabular}{l l l l}
\hline
Optimiser & SGD, mom.\ $0.9$ & $E_{\mathrm{base}}$ & $5$ \\
Weight decay & $5{\times}10^{-4}$ & $E_{\max}$ & $10$ \\
Grad.\ clip & $1.0$ & $E_{\mathrm{dark}}$ & $2$ \\
Batch & $64$ & $s_e$ & $0.7$ \\
$\eta_0$ & $0.01$ & $R_{\mathrm{intra}}$ & $2$ \\
$\gamma$ & $0.998$ & $\rho$, $\kappa$ & $0.95$, $0.5$ \\
\hline
\end{tabular}
\end{table}

\textit{Baselines.} FedAvg~\cite{mcmahan2017fedavg} averages
visible-orbit models~\eqref{eq:fedavg_global};
FedProx~\cite{li2020fedprox} adds a proximal term
($\mu = 0.01$); APFL~\cite{deng2020apfl} interpolates the
FedAvg global and a local orbit model with an adaptively
learned mixing coefficient; Ditto~\cite{li2021ditto} trains a
per-orbit personal model with an $\ell_2$ proximal term
($\lambda_{\mathrm{Ditto}} = 0.1$) toward a FedAvg global
trained in parallel. The optimiser, batch size, initial
learning rate, and scheduler are shared across methods.

\textit{Evaluation.} For each orbit $l$, $\mathrm{Acc}_l$ is
the accuracy of $\widetilde{\mathbf{w}}_l$ on orbit $l$'s test
partition (the global model for FedAvg and FedProx, the
personalised model for APFL and Ditto,
and~\eqref{eq:m_plane_model} for FedOrbit). The primary metric
is the per-orbit mean $\tfrac{1}{L}\sum_l \mathrm{Acc}_l$,
averaged over the last ten evaluated rounds; the secondary
metric is the per-orbit spread $\max_l \mathrm{Acc}_l -
\min_l \mathrm{Acc}_l$. All methods were implemented in PyTorch using the same
partitions and visibility matrices in each repetition.

The hyperparameters in Table~\ref{tab:hparams} are fixed across
the three datasets and both partitions. Adaptation to
heterogeneity is provided by $\beta$ in~\eqref{eq:m_beta}, not
by dataset-specific tuning. A broader sensitivity analysis of
$\kappa$, $\rho$, and $s_e$ is left for future work.

\section{Results and Analysis}
\label{sec:results}

Table~\ref{tab:res_main} reports the per-orbit personalised
accuracy and the per-orbit accuracy spread on all three
datasets under both partitions. FedOrbit achieves the highest
accuracy in five of six settings and is within $0.9$
percentage points of the best result in the sixth.

\begin{table}[t]
\centering
\caption{Personalised accuracy (top) and per-orbit spread (bottom)}
\label{tab:res_main}
\renewcommand{\arraystretch}{1.05}
\setlength{\tabcolsep}{3.5pt}
\begin{tabular}{l c c c c c c}
\hline
& \multicolumn{2}{c}{\textbf{EuroSAT}}
& \multicolumn{2}{c}{\textbf{So2Sat}}
& \multicolumn{2}{c}{\textbf{RESISC45}} \\
\textbf{Method} & Dir. & Path. & Dir. & Path. & Dir. & Path. \\
\hline
\multicolumn{7}{l}{\textit{Accuracy}}\\
FedAvg   & 68.7 & 20.0 & 38.9 & 16.9 & 37.3 & 14.8 \\
FedProx  & 68.5 & 20.3 & 39.1 & 16.9 & 38.1 & 15.4 \\
APFL     & 60.9 & 74.7 & 39.9 & 40.7 & 32.3 & 54.4 \\
Ditto    & 58.5 & \textbf{95.8} & 35.7 & 84.0 & 36.3 & 77.7 \\
\textbf{FedOrbit} & \textbf{76.4} & 94.9 & \textbf{50.9} & \textbf{86.9} & \textbf{54.2} & \textbf{86.3} \\
\hline
\multicolumn{7}{l}{\textit{Spread}}\\
FedAvg   & 28.4 & 78.1 & 26.1 & 68.2 & 10.0 & 53.3 \\
FedProx  & 31.0 & 78.2 & 27.9 & 68.4 &  9.9 & 55.4 \\
APFL     & 40.2 & 53.9 & 31.6 & 70.2 & 12.1 & 38.4 \\
Ditto    & 49.0 & \textbf{9.7} & 32.6 & 17.6 & 11.6 & 17.6 \\
\textbf{FedOrbit} & \textbf{15.8} & 12.5 & \textbf{24.7} & \textbf{15.8} & \textbf{8.7} & \textbf{8.9} \\
\hline
\end{tabular}
\end{table}

\textit{Pathological partition.} With disjoint inter-orbit
classes, FedAvg and FedProx achieve only $14.8\%$ to $20.0\%$
because the global classifier serves only the classes on
frequently visible orbits. APFL improves through interpolation
but is regularised toward the same biased anchor. Ditto is the
strongest baseline; FedOrbit gains $8.6$ percentage points over
Ditto on RESISC45 and $2.9$ on So2Sat, while trailing Ditto by
$0.9$ percentage points on EuroSAT.

\textit{Dirichlet partition.} A different pattern is observed.
Ditto is $37$ to $48$ percentage points lower than under
pathological partitioning, and
APFL also degrades because strong personalisation is less
beneficial when orbit distributions overlap. FedOrbit gains $7.7$,
$11.0$, and $16.1$ percentage points over the strongest baseline on
EuroSAT, So2Sat, and RESISC45. The adaptive feature
decomposition provides a consistent explanation:
$\beta$~\eqref{eq:m_beta} is moderate under Dirichlet
partitioning ($\approx 0.44$) and small under pathological
partitioning ($\approx 0.11$). The personal feature extractor
therefore tracks the global average more quickly under Dirichlet
partitioning and retains orbit-specific information for longer
under pathological partitioning.

\textit{Per-orbit fairness.} The spread block of
Table~\ref{tab:res_main} operationalises the fairness term
in~\eqref{eq:objective}. FedOrbit has the smallest spread in
five of six settings and is within $3$ percentage points of the
smallest in the sixth (EuroSAT pathological). Under pathological
partitioning, the global baselines leave spreads of $53$ to $78$
percentage points, confirming poor accuracy for low-visibility
orbits.

\textit{Component interpretation.} As the mechanisms operate
jointly, Table~\ref{tab:res_main} does not isolate their
individual effects. Fixed hyperparameters avoid
dataset-specific tuning, while the change in $\beta$ is
consistent with stronger sharing under Dirichlet partitioning
and greater orbit-specific retention under pathological
partitioning. A controlled ablation is left for future work.

\section{Conclusion}
\label{sec:conclusion}

FedOrbit addresses the coupling between orbit-level data
heterogeneity and visibility-dependent participation in LEO
constellations. It combines continuous orbit-level training,
class-aware hierarchical aggregation, quality-weighted feature
aggregation with return-rate dampening, and adaptive feature
decomposition. Across three remote-sensing datasets,
FedOrbit achieves the highest accuracy in five of six settings
and is within $0.9$ percentage points of the best result in the
sixth. It also produces the smallest per-orbit accuracy spread
in five of six settings. Future work will evaluate individual
component contributions, hardware deployment, and larger
dynamic constellations.

\bibliographystyle{ieeetr}
\bibliography{Ref}
\end{document}